\documentclass[
twocolumn,
]{ceurart}

\usepackage{subcaption}
\usepackage{listings}
\usepackage{booktabs}
\usepackage{subfiles}
\begin{document}
\definecolor{vic-color}{rgb}{0.158, 0.188, 0.878}
\newcommand{\Vic}[1]{\textcolor{vic-color}{$_{Vic}$[#1]}}
\newcommand{\kmy}[1]{\textcolor{purple}{$_{min}$[#1]}}


\copyrightyear{2023}
\copyrightclause{Copyright for this paper by its authors. Use permitted under Creative Commons License Attribution 4.0 International (CC BY 4.0).}
\conference{Fifth Knowledge-aware and Conversational Recommender Systems (KaRS) Workshop @ RecSys 2023, September 18--22 2023, Singapore.}
\title{A Conversation is Worth A Thousand Recommendations: \\ A Survey of Holistic Conversational Recommender Systems}


\author[12]{Chuang Li}[%
email=lichuang@u.nus.edu,
orcid=0009-0006-8112-3505,
url=https://github.com/lichuangnus/CRS-Paper-List
]
\cormark[1]
\author[1]{Hengchang Hu}[
email=hengchang.hu@u.nus.edu,
orcid=0000-0001-7847-0641
]

\address[1]{National University of Singapore, Singapore}
\address[2]{NUS Graduate School for Integrative Sciences and Engineering, Singapore}

\author[1]{Yan Zhang}[%
email=eleyanz@nus.edu.sg,
orcid=0000-0002-5336-7100
]

\author[1]{Min-Yen Kan}[%
email=kanmy@comp.nus.edu.sg,
orcid=0000-0001-8507-3716
]
\address[3]{The Chinese University of Hong Kong, Shenzhen, China}

\author[13]{Haizhou Li}[
email=haizhou.li@nus.edu.sg,
orcid=0000-0001-9158-9401
]

\cortext[1]{Corresponding author.}

\begin{abstract}
Conversational recommender systems (CRS) generate recommendations through an interactive process. 
However, not all CRS approaches use human conversations as their source of interaction data; the majority of prior CRS work simulates interactions by exchanging entity-level information. As a result, claims of prior CRS work do not generalise to real-world settings where conversations take unexpected turns, or where conversational and intent understanding is not perfect.   To tackle this challenge, the research community has started to examine {\it holistic CRS}, which are trained using conversational data collected from real-world scenarios.  Despite their emergence, such holistic approaches are under-explored.

We present a comprehensive survey of holistic CRS methods by summarizing the literature in a structured manner. 
Our survey recognises holistic CRS approaches as having three components: 1) a backbone language model, the optional use of 2) external knowledge, and/or 3) external guidance.  We also give a detailed analysis of CRS datasets and evaluation methods in real application scenarios.
We offer our insight as to the current challenges of holistic CRS and possible future trends. 
\end{abstract}


\maketitle

\section{Introduction}

Conversational Recommender Systems (CRS) integrate conversational and recommendation system technologies, to facilitate users in achieving recommendation-related goals through conversational interactions \cite{A-survey-jannach}.
In contrast to traditional recommendation systems, which act in a single (one-shot) round of interaction, CRS support multiple rounds of interaction, allowing the system to make multiple attempts in recommendation.

\begin{figure*}[ht]
     \centering
     \includegraphics[width=\linewidth]{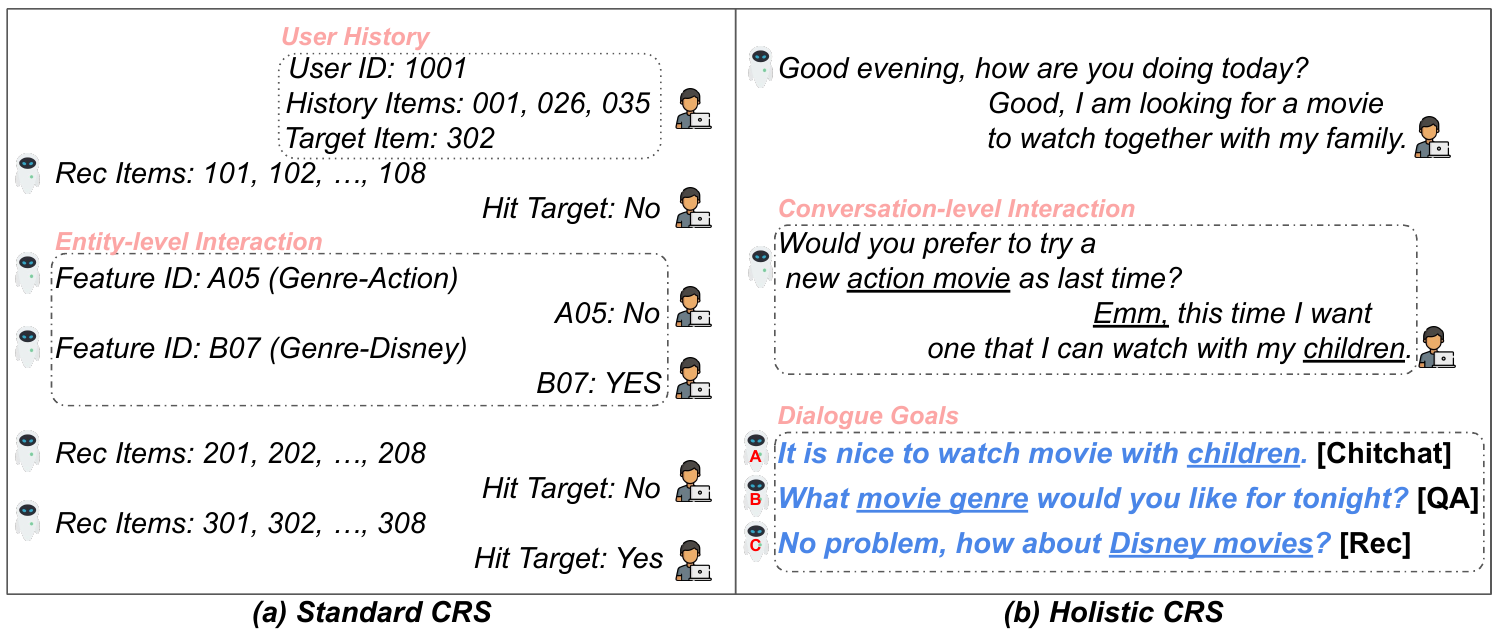}
    \caption{Examples of standard and holistic CRS. a) Standard CRS support multi-round interaction only at the entity level b) Holistic CRS support multi-round and multi-goal interaction at the conversation level. }
        \label{fig:example}
\end{figure*}

In much prior work on CRS, the multiple rounds of interaction are simulated by entity-level interaction, consisting of a sequence of entity-level features \cite{SAUR, EAR}. For example in Figure~\ref{fig:example}(a), the entity-level interaction process is illustrated by how the system selects the ``Feature ID'' of \textit{<Genre-Disney>} from its feature list, and the simulated human response of \textit{<Yes>} will be directly returned to the system.  Such a framing of the CRS task focuses on recommendation and decision-making strategies, which neglect the conversational element, such as possible inaccuracies in understanding the human language that makes up the conversation.  Inaccurate conversation comprehension, gauging of intent and incorrect response generation \cite{DST1, DST} as well as information inconsistency \cite{consistency} are a regular occurrence in human conversation, yet much research on CRS have simply abstracted away from these defining characteristics.  This is due to its presumption that the entity-level interaction is invariably accurate \cite{EAR}. As a result, the application and evaluation of such systems in real-world situations pose significant challenges.

Thus there is a dichotomy in CRS research.  Most 
CRS do not assume actual human conversations for interaction, only simulating the interaction with entity-level information \cite{gao_advances_2021, EAR}.
However, there are also prior work that relax this constraint and tackle conversational recommendation based on actual human conversations \cite{li_towards_2018_ReDial, liu_towards_2020_DuRecDial}.  Besides recommendation and decision strategy, these works also tackle the aforementioned conversational challenges in language understanding, generation, topic/goal planning and knowledge engagement.  To distinguish these two forms of CRS research, we divide the current research works in CRS into \textbf{\textit{standard CRS}} (the former, more prevalent form of prior CRS work), and what we term \textbf{\textit{holistic CRS}} (which assumes a wider scoping of the CRS task) based on the input and output formats, as shown in Figure~\ref{fig:hier}.



Research on holistic CRS is burgeoning, and it is timely to comprehensively survey such works to better organise and make sense of their contributions and gauge their potential future directions. 
This is needed to effectively utilize holistic CRS and the conversational datasets collected from real-world scenarios  \cite{christakopoulou_towards_2016,li_towards_2018_ReDial} that train them, in practical contexts. 
Holistic CRS adopt real, conversation-level interaction and target multiple dialogue goals, as shown in Figure~\ref{fig:example}. Given the same entity pair <Genre-Disney> as the standard CRS in subfigure (a), the holistic system in subfigure (b) must generate questions like \textit{``What movie genre would you like for tonight?''} and understand its related response correctly, before they use \textit{<Genre-Disney>} for the recommendation. For the same question, the user may give unexpected answers like \textit{``Show me a new movie this year!''}, inconsistent with the movie genre. Moreover, holistic CRS is required to leverage the rich contextual information inferred from the conversations \cite{semantic} and from the semantic context. For example, given the input \textit{``Emm''} in the user's second response, a holistic CRS might infer that the previous recommendation was unsatisfactory, prompting it to make a new and different recommendation. 

The main challenges in the task of a holistic CRS are thus ones such as the following: {\it How to understand the users' intentions with limited contextual information? How should we generate reasonable responses with high recommendation quality?  When faced with different inferred conversation goals, which goal should be pursued now?}

We systematically analyse the current holistic CRS work solving the above problems (\S\ref{Methods3}), decomposing them into three components: 1) a backbone language model, and optional components incorporating 2) external knowledge and 3) external guidance.  We follow this with an analysis of the datasets (\S\ref{datasets4}) and evaluation methods (\S\ref{evaluation}). We investigate the key challenges and promising research trends in this area (\S\ref{future6}). 
To the best of our knowledge, this is the first survey on CRS with a special focus on conversational (``holistic'') approaches. Our contributions are: 

\begin{enumerate} 
    \item We provide a clear landscape of the tasks, models and hierarchical structure of holistic CRS.
    \item We summarise, analyze and critique the existing methods, datasets and evaluation methods for selected works in a well-structured manner.
    \item We outline key challenges, constraints and future directions for holistic CRS. 
\end{enumerate}

\section{Definition and Background}\label{background2}

\begin{figure}[h]
  \includegraphics[width=\linewidth]{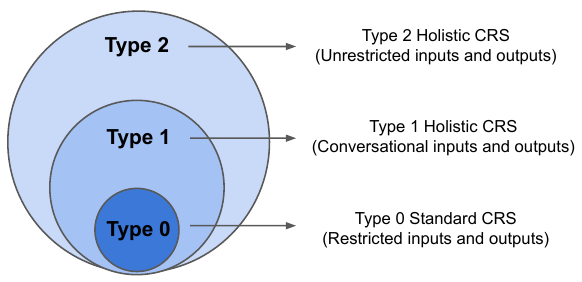}
  \caption{Hierarchical structure of CRS in terms of input and output types}
  \label{fig:hier}
\end{figure}

In Figure~\ref{fig:hier}, we split the field of CRS research into two distinct branches: standard and holistic CRS, further delineating them into Types 0, 1, and 2, based on their input--output dynamics.  \\

\textit{\textbf{Type 0 standard CRS}, limited to entity-level inputs and outputs, is restricted in scope of interaction; e.g., \cite{SAUR,EAR}.} 

\textit{\textbf{Type 1 holistic CRS} takes conversation as input and yields either entity-level recommendations or conversational responses, encompassing query interpretation and tailored linguistic outputs; e.g., \cite{li_towards_2018_ReDial, INSPIRED-shirley}.} 

\textit{\textbf{Type 2 holistic CRS} is more expansive, accepting and producing unrestricted inputs--outputs formats including conversations, knowledge and multimedia; e.g.,  \cite{multimodal1, multimodal2}.} \\

 Holistic CRS differ from standard CRS approaches in the following aspects: 1) The final goal for holistic CRS is to guide or convince users to accept the recommendation through multi-rounds of conversations. 
 2) Holistic CRS start from the conversations and ends by generating either recommendation results or responses.  
 3) Holistic CRS methods are evaluated on both recommendation and language quality using both automatic and human evaluation measures.

\begin{figure*}[th]
  \includegraphics[width= \linewidth]{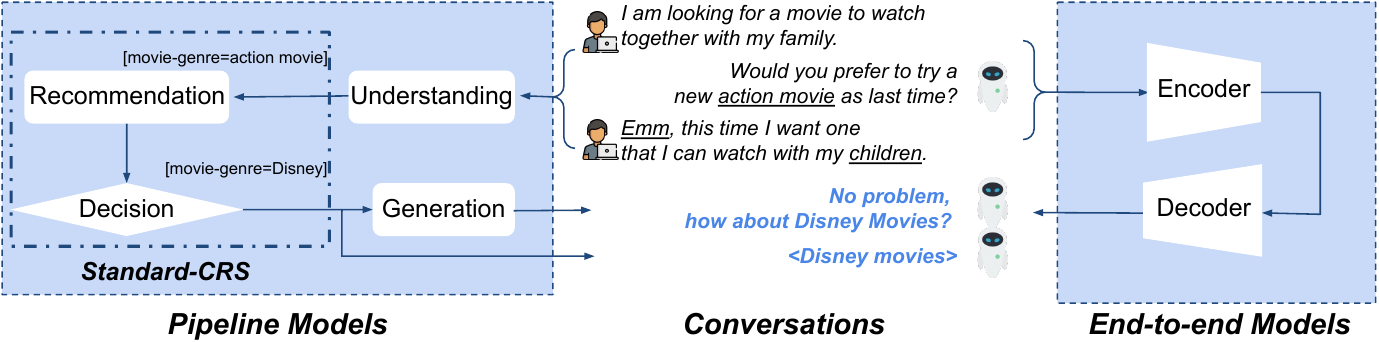}
  \caption{Pipeline models and end-to-end models in holistic CRS. Left: Pipeline models for holistic CRS  include understanding, recommendation, decision and generation units while standard CRS only contain recommendation and decision units. Right: End-to-end holistic CRS with an encoder--decoder structure.}
  \label{fig:models}
\end{figure*}

\subsection{Task Definition}

In a task-oriented dialogue system, we restrict our consideration to the scenario where a singular system interacts with one individual user, denoted by $u$, and pre-determined items, represented by $i$.
Each dialogue contains $T$ turns of conversations, denoted as $C = $\{$s_j^{system}, s_j^{user} $\}$^T_{j=1}$, where each turn contains a single turn from the system and its associated response from the user. The user's entity-level interaction history of past $j$-th turn is denoted as $E^u_j =$\{$i_1^{(u)}, ..., i_j^{(u)}$\} and dialogue history with past $j$-th turns is denoted $C^u_j =$\{$[s_1^{system}, s_1^{user}], ..., [s_j^{system}, s_j^{user}]$\}. Some methods provide knowledge or external guidance, which we denote as $K$.
The target function for holistic CRS is expressed in two parts: to generate 1) next item prediction $i_{j+1}$ and 2) next system response $s_{j+1}^{system}$. In summary, at the $j$-th turn, given the user's interaction history and contextual history, CRS generates either an entity-level recommendation results $i_{j+1}$ or a conversation-level system response $s_{j+1}^{system}$, shown in Formula~\ref{function}. 

\begin{equation}
\label{function}
y^* = \prod_{j=1}^T P_\theta~(i_{j+1}, s_{j+1}^{system}|E_j^u, C_j^u, K ) 
\end{equation}


\subsection{Structure of CRS}

Figure \ref{fig:models} shows the two prevalent model structures in holistic CRS, which are pipeline and end-to-end models. 

The pipeline structure of CRS contain four parts: \textit{understanding unit, recommendation unit, decision unit} and \textit{generation unit}. The understanding unit takes in dialogue and converts them into an entity--value pair for the recommendation unit to generate possible entity outputs. The decision unit controls the dialogue flow, while the generation unit generates the response accordingly. 

With the development of Hierarchical Recurrent Encoder--Decoder (HRED) structure    \cite{li_towards_2018_ReDial} and transformer-based encoder--decoder structure \cite{penha_what_2020}, components such as the understanding, decision and generation units are merged together to form an end-to-end structure. 


\section{Ontology and Existing Surveys}

We aim to conduct an exhaustive survey on holistic CRS, focusing on Types~1 and 2 of our hierarchy. Our primary sources comprise leading NLP and Information Retrieval (IR) conferences and journals, as exemplified by premier venues such as
ACL, ACM, AAAI and ScienceDirect.  Furthermore, we delve into publicly accessible online resources, filtering papers by all variants of search terms in ``conversational recommender systems''. Matching work are then refined 
based on the following three criteria, regarding the features of the presented work:
1) It supports conversations as an input type.
2) It provides recommendation responses at either entity or conversation levels.
3) It facilitates multi-round interactions.
For each selected work, we focus on the methodologies, datasets, and evaluation metrics.

While there exist surveys that offer an all-encompassing view of CRS, encompassing both standard and holistic CRS \cite{A-survey-jannach,gao_advances_2021, radlinski2022subjective, jannach2023evaluatingcrs}, our survey purposefully structured and limited in scope to illuminate the evolution and development of holistic CRS only, particularly in their handling of conversational data. Works centred on Type 0 standard CRS, given their lack of conversational aspects, are intentionally omitted.


\begin{figure}[t]
  \includegraphics[width=\linewidth]{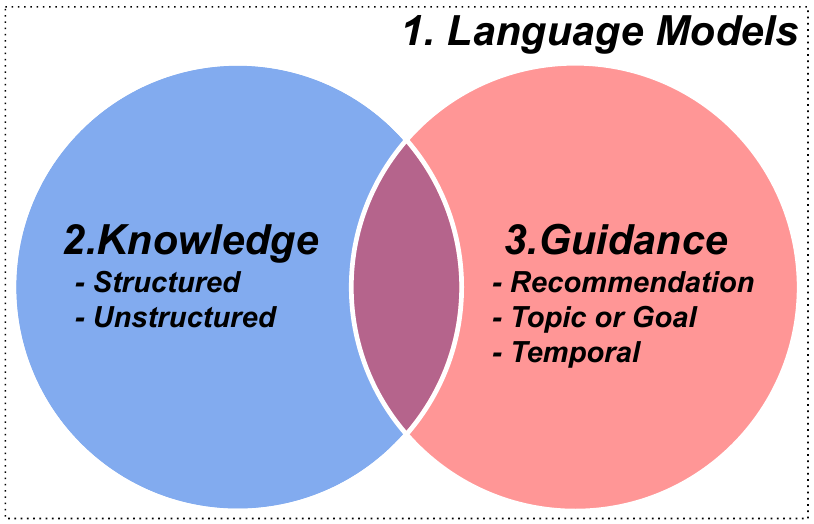}
  \caption{Components of holistic (Type~3) Conversational Recommendation System approaches: 1) requisite backbone language models, and optional components incorporating 2) external knowledge and/or 3) external guidance}
  \label{fig:approach}
\end{figure}

\section{ Main Approaches \& Discussion} \label{Methods3}

Current holistic CRS approaches are primarily structured around three main components, as illustrated in Figure~\ref{fig:approach}: 1) Language Models (LMs); 2) Knowledge; and 3) Guidance. 
A majority of holistic CRS systems hinge on LMs (\S\ref{LM}), encompassing machine learning, deep learning, and pre-trained language models (PLMs), for foundational dialogue operations. However, these LMs often fall short in recommendation and commonsense reasoning. 
To bridge this gap, additional external knowledge (\S\ref{knowledge}) and guidance (\S\ref{guidance}) are integrated, either independently or jointly. 
This section delineates the evolutionary path of their development, offering insights into their limitations and potential avenues for future progress.


\subsection{Language Models} \label{LM}

LMs serve as the backbone for holistic CRS in recommendation response generation with the evolution from machine learning  \cite{christakopoulou_towards_2016}, deep learning  \cite{li_towards_2018_ReDial, recommendation-as-a-communication_GoRecDial} to PLMs  \cite{penha_what_2020, INSPIRED-shirley, manzoor2022inspired2}. The most popular LMs for response generation are HRED-based sequential models and transformer-based PLMs. These language models  adopt a framework of end-to-end training, enabling them to be simultaneously trained in both conversation and recommendation tasks \cite{li_towards_2018_ReDial, recommendation-as-a-communication_GoRecDial}. 
 
Recent advancements in natural language processing (NLP) highlight the efficacy of PLMs like BERT and GPT~\cite{BERT, zhang2019dialogpt} in language generation and commonsense reasoning. 
Although those PLMs are not inherently optimized for CRS, researchers have explored their capabilities for holistic CRS tasks like recommendations and response generation. 
\citeauthor{penha_what_2020} evaluated BERT's innate ability for recommendations using text-format probes for item or genre predictions without fine-tuning. In another line of work, \citeauthor{INSPIRED-shirley} enhanced conversational tasks by adapting PLMs to produce varied recommendation responses incorporating social strategies, like encouragement or persuasion \cite{INSPIRED-shirley, manzoor2022inspired2}.
Taking a multifaceted approach, \citeauthor{uniCRS} segmented recommendation response generation into multiple tasks, including goal or topic planning, item recommendation and response generation. While having distinct tasks, they pre-trained a PLM end-to-end, underscoring the connection between holistic CRS and LMs and validating the effectiveness of the end-to-end training paradigm. \\

\textbf{\textit{Discussion}}. \textit{While PLMs can generate context-specific recommendation responses, they often fall short of meeting the dual requirements of recommendation accuracy and language quality, resulting from the phases of 1) pre-training and 2) online training.}

\textit{The inherent limitation of PLMs stems from their design for universal application.  In contrast, recommendation tasks are focused and specific to certain domains \cite{li_towards_2018_ReDial, job-rec}. The implicit knowledge derived from general pre-training is insufficient to support them in making high-quality recommendations.  Pre-training LMs with explicit task-specific knowledge is a solution, but comes associated with high costs and complications \cite{P5, uniCRS}. Transferring such knowledge across diverse domains or user groups for real-world applications still poses a considerable challenge.}
  
\textit{Holistic CRS rely heavily on online training, enabled by conversational interactions with benchmark datasets (\S\ref{datasets4}). However, the restricted knowledge available in those datasets poses a formidable challenge for PLMs to generate quality recommendation responses, necessitating a model capable of integrating additional knowledge or guidance to facilitate preference tracking and response generation.  }


\subsection{External Knowledge} \label{knowledge}

Inherent limitations regarding implicit knowledge stored in PLMs are addressed in holistic CRS by integrating external knowledge. This enhances their capabilities in prediction, reasoning, and explanation. Methods augmented with knowledge often utilize graph convolutional networks (GCNs) \cite{GCN} or relational graph convolutional networks (R-GCNs) \cite{RGCN} to extract knowledge representation from structured sources like knowledge graphs (KGs), or unstructured ones such as reviews. This representation is then incorporated into PLMs through semantic alignment or knowledge fusion techniques, enabling the production of refined recommendations \cite{kang_self-attentive_2018,zhang_kecrs_2021,zhou_c2-crs_2022}. We now delve into holistic CRS approaches that leverage both structured and unstructured knowledge sources.

\subsubsection{Structured knowledge } \label{structure}

Knowledge Graphs (KGs) are a prevalent source of structured knowledge.  However, to be employed for holistic CRS tasks, they need to be transformed into an appropriate representation before the knowledge and textual features can be integrated.
 
KGs are typically represented by triplets comprising entities and relationships; e.g., \textit{<Movie A-Genre-Disney>} where nodes representing item entities (\textit{Movie A}) are connected to non-item entities (\textit{Disney}) via edges that indicate relationships (\textit{Genre}). In knowledge-enhanced CRS, the entities mentioned in conversations are first matched with entities in external KGs. Subsequently, graph propagation is performed to encode the KG's structural and relational information into knowledge representations \cite{chen_towards_2019}. Techniques like GCN and RGCN are employed in this stage to recurrently update node representations based on their neighbouring nodes.
With the obtained knowledge representations, there are two main research directions in applying KGs to holistic CRS, which we denote as 1) node-level entity prediction and 2) edge-level path reasoning  \cite{open-dial-KG}.

 \textbf{Node-level entity prediction} in holistic CRS enhances response generation by incorporating additional item entities from the KG \cite{chen_towards_2019, RecInDial}. In this usage, LMs extract knowledge representations from the KG and convert them into item-specific vocabularies, which are then integrated into recommendation responses. As a result, such responses are more fluent and informative, aligning closely to the original conversations and consistent with the user's interests  \cite{chen_towards_2019,RecInDial, kers-knowledge}. 

 \textbf{Edge-level path reasoning} provides a better approach to interpret users' preferences and dynamic shift in interests through the knowledge presentation than node-level entities \cite{zhou_crfr_2021, CR-walker, open-dial-KG, xu2020user}.
  A strict, 2-hop KG reasoning is first proposed to interpret the user's preference through two steps (e.g.,\textit{Movie A $\Rightarrow$Actor1$\Rightarrow$Movie B}). For instance, given the user's watching history of Movies~A and B, the model can infer the user's preference for Actor~1 and subsequently confirm its inference through conversation. However, due to the rule-based setting, 2-hop reasoning works well only when users have clearly-defined and straightforward preferences \cite{CR-walker}. In situations where users demonstrate shifting interests, a multi-hop or tree-structure reasoning method is more suitable, translating implicit preference paths in KGs to explicit explanations in dialogues \cite{zhou_crfr_2021, xu-etal-2020-user, TERA}. 
  
Well-constructed KGs enhance comprehensive knowledge representation in entity-level item selection and conversation-level preference reasoning or interpretation \cite{open-dial-KG, TERA}. However, due to the static nature of KGs, inferring the latest features of an item from structured knowledge sources poses significant challenges.


\subsubsection{Unstructured knowledge} \label{unstructure}

In unstructured knowledge sources (e.g., reviews or documents), a text retriever is employed to extract relevant textual segments from external documents. These segments are subsequently either transformed into nodes or edges of a new KG or merged into an existing KG \cite{liao_deep_2019, zhou_c2-crs_2022, lu_revcore_2021,  li-etal-2022-knowledge_plug, yang-etal-2022-improving-meta-crs}. The resultant KG can then be transferred into knowledge representations \cite{li-etal-2022-knowledge_plug, yang-etal-2022-improving-meta-crs, zhang2023variational}.
This method allows unstructured knowledge to supplement static knowledge graphs with contemporary information, allowing holistic CRS to be more versatile.

\textbf{Knowledge Fusion} and \textbf{Semantic Alignment} serve as the primary strategies to bridge the entity and semantic spaces in graph reasoning, leveraging both structured and unstructured knowledge resources. Knowledge Fusion integrates graph embeddings from KGs with text embeddings from LMs, enhancing both entity recommendations and conversational preference interpretations \cite{chen_towards_2019, zhang_kecrs_2021}. 
Recently, \citeauthor{zhou_improving_2020} demonstrate a method that surpasses the performance of current fusion methods for entities and dialogues. They address the semantic gap between conversations and external knowledge with fine-grained semantic alignment techniques that align word-level semantic graphs with entity-level KGs \cite{zhou_improving_2020,wu2023semantic, DICR}. Similarly, for models utilizing unstructured knowledge bases, contrastive learning strategies bridge the semantic gap across embeddings in dialogues, KGs and document reviews, potentially leveraging a spectrum of such knowledge resources \cite{zhang_kecrs_2021}. \\

\textbf{\textit{Discussion}}. \textit{The existing knowledge sources for holistic CRS are constrained in item space. However, as LMs become more robust, the reliance on conventional knowledge sources might decrease, while the necessity for guidance in other modalities may increase. Specifically, specialized knowledge (such as user profile representation and user--item relationship extraction) is likely to become crucial.}

\textit{The advent of powerful large language models (LLMs) serving as LMs, reduces reliance on external knowledge sources. This potentially makes the use of external sources redundant \cite{knowledge, he2023large}. The integration of external knowledge within LMs should start by evaluating a model's capabilities before knowledge incorporation, such as examining the capability of PLM in processing content-based recommendations \cite{knowledge, liu_graph-grounded_2022}.  Recognizing the limitations of LMs before introducing the appropriate knowledge sources is a key issue in the advancement of holistic CRS.}

\begin{table*}[ht]
\small
\tabcolsep=1.75mm
\centering
\begin{tabular}{l|r|rrrr|cc|crr}
\toprule
\textbf{Dataset} & \textbf{\#~P} & \textbf{\#~C}  &\textbf{\#~T} & \textbf{\#~I} & \textbf{\#~M}  &\textbf{Domain} & \textbf{Language} &\textbf{IR} & \textbf{\#~Pos} & \textbf{\#~Neg}\\
\midrule
\textbf{REDIAL} & {20} & {10,006}  &{182,150} & 6,223 & 8.50 & {Movie} & {Freestyle} & 0.96& 94,150 & 15,377 \\
\textbf{TG-ReDial*} & {5} & 8,495  & 109,892 & 11,447 & 2.22 &Movie & Topic-guided& 0.40 & 89,693 & 2,971   \\
\textbf{DuRecDial*} & {4} & 5,678  & 87,301 & 531 & 15.15 &Movie+Music & Multi-type&  0.48 & 47,547 & 14,217  \\ 
\textbf{INSPIRED} & {4} & 801  & 16,982 & 1,378 & 10.05 & Movie & Social strategy & 0.35 & 12,589 & 2,395 \\
\textbf{OpenDialKG} & {3} & 13,802  & 91,209 & 4,232 & 82.49 & Movie+Book & Knowledge path& 0.38 & 63,856 &15,798 \\
 
\textbf{GoRecDial} & {2} & 8,209  & 16,743 & 1,532 & 20.44 & Movie & Game-play&  0.77 & 88,601 & 19,354 \\ 

\textbf{MultiWOZ} & {2}  & 8,420  & 221,588 & 1,737& 238.58 & 7 Other Domains & Multi-domain&  0.68 & 75,732 &26,724 \\
\bottomrule
\end{tabular}

\caption{Statistical analysis of the datasets in holistic CRS research. \#~P, \#~C, \#~T, \#~I, \#~M, \#~Pos, and \#~Neg stand for the number of papers, conversations,  single turns, items, mentions for each item, positive and negative single turns in training data, IR: Informative turns rate. $^*$ Datasets are originally collected in Mandarin Chinese. }
\label{Dataset}
\end{table*}

\subsection{External Guidance}\label{guidance}

Holistic CRS using external guidance train models for supplementary tasks --- inclusive of recommendation, topic/goal planning, and temporal feature representation --- in contrast to knowledge-enhanced models which fuse knowledge into PLMs.
Results from these tasks serve as auxiliary guidance for LMs during recommendation response generation. Some models align both external knowledge and guidance, adopting a hybrid strategy that capitalizes on both dimensions for more robust response generation.

\textbf{Recommendation guidance} utilises approaches akin to template-based generation methods, decoupling conversation and recommendation result generation. LMs are conditioned to separately produce dialogues with placeholders that align with the original context and suggested items or attributes consistent with the user's history \cite{manzoor_generation-based_2021-1,manzoor_towards_2022, uniCRS1, RecInDial, liang-etal-2021-NTRD}. These placeholders are later substituted with corresponding recommendations.

\textbf{Topic or goal guidance} enhances the LM's proficiency in topic or goal planning. Although reinforcement learning techniques are predominantly employed in traditional CRS for action or goal planning, they are challenging to adapt as a representation for LMs \cite{EAR, recommendation-as-a-communication_GoRecDial, uniCRS}.

\textbf{Topic-guided systems} initiate by building topic graphs, capturing or predicting specific target topics like ``action movie'' or ``Disney movie''. LMs subsequently use these graphs to guide recommendation response generation \cite{zhou_towards_2020_TGRedial, topic-guided-CRS-in-multiple-domain, kers-knowledge}.
\textbf{Goal-guided systems} create hierarchical goal-type graphs derived from existing KGs and dialogues. The goal-planning module of the LMs is then trained on diverse dialogue goals, encompassing ``QA'', ``recommendation'', ``greeting'' or ``chitchat''  \cite{liu_towards_2020_DuRecDial,liu_graph-grounded_2022,target-guided, uniCRS}. These objectives also influence the dialogue policy and decision-making processes within holistic CRS.

\textbf{Temporal guidance in CRS} incorporates temporal features to formulate a time-aware representation, emphasizing the explicit and dynamic shift in users' preferences  \cite{zeng_dynamic_2020,improving-sequential}. 
Unlike traditional sequential recommendation systems that have access to users' historical profiles, holistic CRS often lack this depth of historical data. 
To address this gap, temporal features discern between historical dialogue sessions and the ongoing dialogue session, thereby capturing the multifaceted nature of users' preferences \cite{li_user-centric_2022}.
This differentiation allows the modelling of historical user preferences and continues to gather fresh preferences from active interactions. 
Additionally, such features aid in the construction of user profiles based on past behaviours, facilitating the retrieval of similar user profiles based on their relevance, enhancing preference modelling in a time-aware collaborative manner \cite{wang_improving_2022_contextual_time-aware,improving-sequential}.
In a distinct approach, \citeauthor{xu-etal-2020-user} put forth the idea of a user temporal KG, which contains both offline user knowledge in historical conversations and online knowledge in current or future conversation sessions. 
Representing a leap beyond traditional static knowledge graphs, temporal KGs have garnered significant interest \cite{wang_improving_2022_contextual_time-aware, xu-etal-2020-user}. 
In the context of holistic CRS, dynamic reasoning utilizing temporal KGs represents an innovative and burgeoning research domain \cite{xu-etal-2020-user, BERT4REC, TERA, DICR}. \\

\textbf{\textit{Discussion}}. \textit{Present methodologies for integrating external knowledge or guidance largely involve training LMs to interpret fed knowledge or representation, rather than guiding them to independently explore and extract the required information from external resources. This method, akin to ``spoon-feeding'' LMs with knowledge or guidance, contrasts with the envisioned future for holistic CRS. In our view, LMs should be provided with a knowledge ``buffet'', empowering autonomous gathering of necessary information and prioritising reasoning over interpretation \cite{buffet}.}

\begin{table*}[h]
\small
\centering

\begin{tabular}{lc|lc|lc}
\toprule
\multicolumn{2}{c|}{\textbf{Recommendation Accuracy}}&\multicolumn{4}{|c}{\textbf{Language Quality}} \\
\midrule
\textbf{Metrics\hspace{1.5cm}} & \textbf{\#~Papers} &  \textbf{Metrics} & \textbf{\#~Papers} &\textbf{Human Evaluation} & \textbf{\#~Papers} \\
\midrule
\textit{\textbf{Recall@K}} & {19}  &  \textit{\textbf{Distinct-n}}   &  {18}  &  \textit{\textbf{Fluency}}   & {19} \\

 \textit{\textbf{Hit@K}}   & {7}  & \textit{\textbf{BLEU}}   & {15}  & \textit{\textbf{Informativeness}}   & {17} \\
 
\textit{\textbf{NDCG@k}}   & {6} & \textit{\textbf{Perplexity}}   & {9} &  \textit{\textbf{Coherence}}   & {8} \\ 

 \textit{\textbf{MRR@K}}   & {6}  & Knowledge Precision   & {2} &Relevance   & {4}\\ 
 
 F1   & {7}   &Entity Accuracy   & {1}  &Proactivity   & {2} \\
 
 Precision    & {2}  &  Average Entity Number   & {1}  & Knowledge   & {2}\\ 
 
 Turn@K   & {1}   &\textit{\textbf{Topic Consistency}}   & {1}   & Appropriateness   & {2} \\
 
  RMSE & {1}  & \textit{\textbf{Success Rate}}   & {1}   &\textit{\textbf{Consistency}}  & {1}\\ 
 
 
\bottomrule
\end{tabular}
\caption{Evaluation methods in holistic CRS. \#~Papers indicate the volume of work using the associated evaluation method.}
\label{Evaluation}
\end{table*}

\section{Datasets}\label{datasets4}

In the realm of holistic CRS, the interaction between users and systems has led to the collection of several benchmark datasets. While some surveys have primarily summarized data from an item space perspective~\cite{A-survey-jannach}, our focus  is to dive deeper into the publicly-available holistic CRS datasets. Our intention is understand datasets beyond traditional boundaries, expounding specifically on two dimensions: entity information and language quality \cite{li_towards_2018_ReDial,INSPIRED-shirley,zhou_towards_2020_TGRedial,recommendation-as-a-communication_GoRecDial,open-dial-KG,liu-etal-2021-durecdial2,zhou_towards_2020_TGRedial}.

\subsection{Statistical analysis}

Table~\ref{Dataset} presents a statistical analysis of various datasets, detailing each dataset in terms of both entity and linguistic characteristics.
In terms of entity space, the scale of a dataset is measured by the number of conversations and items it contains; while the informativeness is measured by the number of conversation turns and the number of mentions of specific items within them. Interestingly, our analysis reveals that a longer conversation does not necessarily correspond to mentions of more items. Rather we believe that ensuring a consistent frequency of item mentions is paramount for the recommendation system's learning efficacy~\cite{frequency}.

From the perspective of language, most datasets are compiled from predominately English data and focus on the movie domain.
Recent datasets indicate a decline in the ratio of informative turns. This trend aligns with real-world conversational patterns, where interactions are transforming into conversations that contain a growing amount of general or chit-chat content \cite{INSPIRED-shirley, manzoor2022inspired2}. This observation reinforces our belief that an optimal dataset should capture authentic human behaviour and not merely translate entity-centric data into dialogues.
The data also suggests that positive turns --- ones that provide constructive or affirmative feedback --— are more valuable for recommendations compared to negative ones \cite{critique-to-preference, positive_feedback2}.
In sum, it is not merely about the volume of training data, but about the quality, authenticity, and informativeness of the conversations therein.

\subsection{Limitations}
The objective of Holistic CRS datasets is to accurately emulate real-world scenarios and offer labelled information for efficient learning. However, our evaluation reveals three primary limitations in the existing datasets: First, some datasets diverge from real-world conversations, which impedes the quality of learned interactions \cite{recommendation-as-a-communication_GoRecDial}. A notable example is the game setting where the dialogue's objective is to guess a target item, disrupting the natural flow of conversation as seekers are already privy to the target item's identity.
Second, a significant proportion of datasets predominantly focus on the movie domain \cite{li_towards_2018_ReDial,chen_towards_2019}, potentially damaging the generalizability of conclusions drawn on CRS research.
Third, current datasets do not offer sufficient labels outside the confines of the item space \cite{li_towards_2018_ReDial, INSPIRED-shirley}. Addressing these shortcomings will be pivotal for productive future research in holistic CRS.

\section{Evaluation Methods}\label{evaluation}

CRS generate both recommendation results and responses. Their evaluation require appropriate mechanisms to assess the quality of both the recommended  items and the resulting dialogue as a whole. Existing evaluation methods examine both recommendation accuracy (as in traditional recommendation systems) and language quality (as in NLP language modelling) separately, using both metrics and human evaluation. We compile the frequency of these methods from the works in \S4 as Table~\ref{Evaluation}.

\subsection{Recommendation Evaluation}
Recommendation evaluation metrics categorise along three lines: point-wise accuracy methods (RMSE), decision support methods (F1) and ranking-based methods (Recall@K). 
The evaluation metrics for holistic CRS are similar to those in standard CRS, where they mostly evaluate the recommendation from the item level. However, for holistic CRS, it is equally important to evaluate the recommendation performance separately at the conversation level in order to ensure information consistency in response generation \cite{RecInDial}. 

\subsection{Language Evaluation}
While most of the recommendation results can be evaluated with metrics, it still requires human beings to evaluate the language generation quality as the golden standard. Metric-based approaches, as auxiliary solutions, provide a fast and simple evaluation of holistic CRS. Language evaluation metrics such as \textit{Distinct n-gram, BLEU} and \textit{Perplexity} evaluate language quality regarding diversity and fluency.

\textbf{Human evaluation} provides a fair evaluation of different models from the viewpoints of users and in a double-blind way \cite{manzoor_towards_2022,christakopoulou_towards_2016}. It is relatively fast and convenient for human annotators to provide a high-quality evaluation in terms of \textit{fluency} and \textit{informativeness}. However, as the human evaluation may only be limited to one or few turns over the whole conversation, it is challenging for the annotators to fully examine the \textit{coherence} and \textit{consistency}, which generally requires the full understanding of dialogue \cite{consistency}.
  
Unlike recommendation systems which merely compare item rankings with respect to the target item, in holistic CRS, implicit features like personality, persuasion, and encouragement also contribute to the success of a recommendation \cite{INSPIRED-shirley}. Evaluating a system based on user experience remains challenging. It is imperative to introduce automatic assessment methods for both system-generated quality and user-centric experiences. \cite{jannach2023evaluatingcrs, Auto_eval,auto_eval2, evaluation_user_simulator}.

\section{Challenges \& Future Trends}\label{future6}

As we have detailed the development of holistic CRS, we now highlight current challenges and suggest future directions to round out our overview.

\textbf{\textit{Language generation quality and style}}.
Current holistic CRS methods do not meet the requirements for practical application due to their inferior language quality scores in human evaluation, even when compared to retrieval-based methods  \cite{manzoor_retrieval-conf, manzoor_towards_2022, zhang_evaluating_2020}.
Successful recommendation responses need to supplement explicit prediction results by accounting for implicit features like social strategy and language styles (e.g., encouragement and informativeness \cite{INSPIRED-shirley, critique-to-preference, positive_feedback2}). As recommendation outcomes often draw from an external or enriched knowledge structure, future research should focus on 1) elevating language quality to garner positive user feedback \cite{end-to-end}, and 2) emphasizing preferred language styles to enhance user acceptance \cite{style}.

\textbf{\textit{User-centric holistic CRS}}. 
Holistic CRS has made strides towards user-centricity by facilitating conversational feedback between the user and the system. Nonetheless, its feedback and recommendation spectrum is still restricted. To enhance its efficacy, future versions of holistic CRS should prioritize personalised experiences for individual users by harnessing multi-modal data from item categories and user profiles. Moreover, attending to users' personal feedback and latent preferences is key for building a superior user modelling framework, resulting in more pertinent recommendations \cite{pramod2022conversationalsurvey}. Additionally, incorporating other LMs or AI-generated content (AIGC) into recommendation feedback could also be a promising avenue \cite{zero-intent, AIGC}.

\textbf{\textit{Unified model for holistic CRS}}. 
Large Language Models (LLMs) have significantly advanced task-oriented dialogue systems, allowing for integrated handling of various tasks in a conversational manner \cite{GPT3, LamDA}. In the realm of recommendation systems, some research has adopted a two-phase training approach (pre-training and fine-tuning), leveraging text for recommendations, reasoning and explanation \cite{BERT4REC, Hitter:hierachical, li2023text}. 
Yet, while there's a push to integrate PLMs into CRS tasks using a text-to-text paradigm, the broader holistic CRS research domain has not achieved a standardized problem framework, which would enable seamless integration with task-specific models and swift adaptation to similar tasks across different domains \cite{RecInDial, zhou_improving_2020, P5}. LLMs, on their own, cannot address every CRS challenge. Current holistic CRS models lean heavily on complex ensemble architectures that merge LMs with external knowledge or guidance. As such, crafting a unified model framework with consistent problem definitions remains a pivotal research avenue \cite{RecInDial, zhou_improving_2020}.

\section{Conclusion}

Despite the rising interest in standard conversational recommendation systems which are restricted to entity-level input and output, our study reveals the necessity and current negligence of holistic CRS, which encompasses all forms of input and output, catering for real-world situations.
In this paper, we systematically describe the important components of holistic CRS, including 1) language models, 2) knowledge resources, and 3) external guidance. To the best of our knowledge, our survey is the first systematic review specifically dedicated to holistic CRS with conversational approaches, which further summarized common datasets, evaluation methods and challenges.
Existing ascendant works enlighten a number of promising future directions from the above perspectives.
Through clear landscapes in holistic CRS, we hope to attract more attention to explore a more natural and realistic setting in this challenging but promising area.

\bibliography{CRS-survey}

\section{Appendix}


\subsection{Datasets for CRS}
We provide a detailed description of each dataset in Table \ref{Description_dataset}. From the perspective of language, each dataset has a different focus. GoRecDial \cite{recommendation-as-a-communication_GoRecDial} uses a game setting to guide the dialogues while TG-Redial \cite{zhou_towards_2020_TGRedial} uses topic to guide the crowd workers. That guidance are utilized to facilitate the CRS towards the target goals. OpenDialKG \cite{open-dial-KG} pairs each conversation with a corresponding KG path while DuRecDial \cite{liu_towards_2020_DuRecDial} further includes user profile and different goals (QA, chitchat, recommendation). These two datasets provide additional knowledge to item space and they are important for knowledge-enhanced models. INSPIRED \cite{INSPIRED-shirley} emphasizes more on the social strategies in making a successful recommendation with more than half utterances involving a social strategy. MultiWoz \cite{multiwoz} collects mainly human-to-human dialogues in multiple domains. Instead of focusing on target item prediction, these two datasets demonstrate real scenarios that aim for successful acceptance in real life. Other datasets that are not publicly available are not included in this survey \cite{xu2020user, conveRSE-dataset,  august}

\begin{table*}[!t]
\small
\setlength{\belowcaptionskip}{-8cm}
\renewcommand{\arraystretch}{1.5}
\begin{tabular}{p{0.2\linewidth} | p{0.75\linewidth}}
\toprule
\textbf{Dataset} & \textbf{Description} \\
\toprule
\textbf{REDIAL}   \cite{li_towards_2018_ReDial} & First CRS dataset collected from crowd workers using a paired mechanism, where one person acts as a recommender and the other person acts as a movie seeker. Crowd workers are free to generate dialogues that meet the basic quality instructions. \\
\midrule
\textbf{TG-ReDial*}   \cite{zhou_towards_2020_TGRedial} & A Chinese CRS datasets with topic-guided dialogues. Using real watching records of real online users to create different topic threads that further generate conversations.\\
\midrule
\textbf{DuRecDial*}   \cite{liu_towards_2020_DuRecDial, liu-etal-2021-durecdial2} & A bi-lingual CRS datasets with additional annotation of users' profile, dialogue goals(QA, chitchat, recommendation) and knowledge. It is collected in Chinese  with paired mechanisms and translated into the English version. \\ 
\midrule
\textbf{GoRecDial}   \cite{recommendation-as-a-communication_GoRecDial} & A goal-driven CRS dataset where the recommender aims to look for the target items by chatting with the seeker. A pair mechanism is adopted and candidate items are provided for each conversation.\\ 
\midrule
\textbf{OpenDialKG}   \cite{open-dial-KG} & A dialogue dataset on movie and book domain with annotated knowledge graphs and relation paths related to each conversation.\\
\midrule
\textbf{INSPIRED}   \cite{INSPIRED-shirley, manzoor2022inspired2} & First CRS dataset proposed to create dialogues with different social strategies and preference elicitation strategies using the paired mechanism. Crowd workers are asked to finish 3 pre-task personality tests and a post-task survey with demographic questions.  \\ 
\midrule
\textbf{MultiWoz}   \cite{multiwoz} & A large transcript of human-to-human dialogues among 7 domains, eg: hotels, restaurants, attractions, taxis, trains, hospitals, police. It contains a large corpus of multi-domain dialogues with labelled dialogue states.\\ 
\bottomrule
\end{tabular}
\caption{Description of datasets. *Datasets are first collected in Mandarin Chinese.}
\label{Description_dataset}
\end{table*}

\end{document}